\documentclass[10pt,twocolumn,letterpaper]{article}

\usepackage{cvpr}              

\DeclareMathOperator{\sigmoid}{sigmoid}
\usepackage{caption}

\definecolor{cvprblue}{rgb}{0.21,0.49,0.74}

\usepackage[pagebackref,breaklinks,colorlinks,allcolors=cvprblue]{hyperref}
\usepackage{amsmath,amssymb,mathtools}
\usepackage[accsupp]{axessibility} 
\usepackage{graphicx}
\usepackage{booktabs}
\usepackage{tabularx,makecell,siunitx}
\usepackage{xcolor}
\usepackage{tikz}
\usepackage{pgfplots}
\pgfplotsset{compat=1.18}
\usepgfplotslibrary{fillbetween} 
\usepackage{xcolor}

\usetikzlibrary{arrows.meta,calc,intersections,positioning}

\def\paperID{13} 
\def\confName{CVPR PBVS}
\def\confYear{2026}

\title{Single-View Seafloor Recovery from Imaging Sonar\\ via Differentiable Rendering}

\author{
  Sevan Brodjian\and
  Michael Hobley \\
  California Institute of Technology\\
  {\tt\small \{sbrodjia, mahobley, perona\}@caltech.edu} \and
  Pietro Perona\\
}
\begin{document}
\maketitle
\begin{abstract}
Sonar is often the only modality suitable for high-resolution imaging underwater due to light attenuation and turbidity. Forward-looking imaging sonar provides measurements over range and horizontal angle but collapses vertical structure into a flat image, creating ambiguities that make 3D recovery challenging. A common use case for imaging sonar is underwater terrain mapping (bathymetry), yet current methods require many views, expensive multi-sensor setups, or significant training data, which limits use and adaptability to new environments. 

We present a training-free method that recovers bathymetry from a single sonar image in under 30 seconds via differentiable rendering, conditioned on a known seafloor tilt. To our knowledge, this is the first differentiable rendering approach for single-view height recovery in sonar. Our method implements differentiable sonar ray tracing and optimizes an explicit height field to reproduce the target image. On synthetic datasets, our approach outperforms a supervised CNN under distribution shift and remains close on rough terrain, while the CNN wins in-distribution. By modeling physically grounded priors of the sonar process, our method adapts across sensor configurations and environments without training data.
\end{abstract}

\begin{figure}[t]
  \centering
   \includegraphics[width=1.0\linewidth]{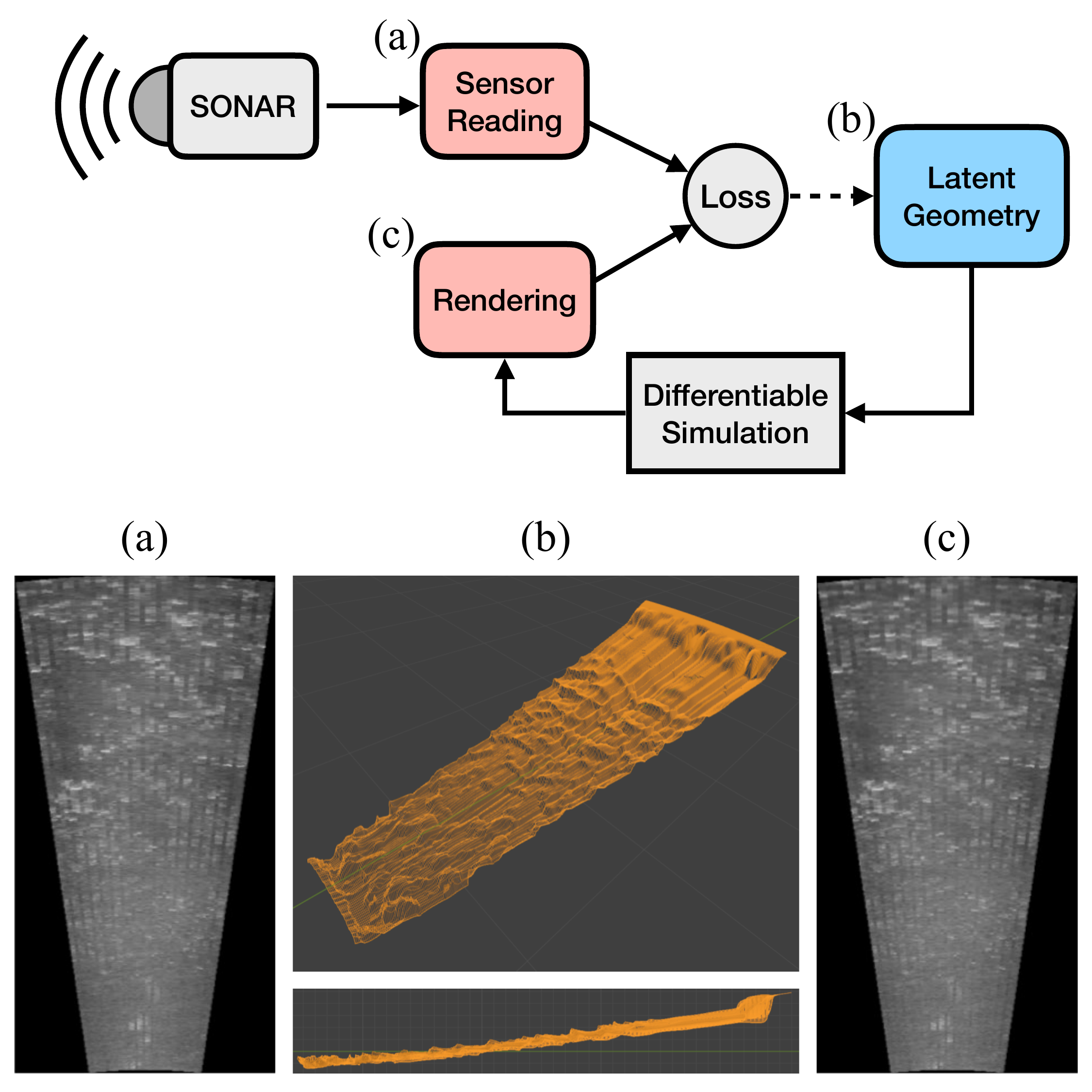}
   \caption{We introduce a differentiable physics simulation of the sonar process which matches a target sensor reading (a) by optimizing a latent height field (b) to minimize the reconstruction error of the simulated output (c) via gradient descent.}
   \label{fig:reconstruction_teaser}
\end{figure}
\section{Introduction}
\label{sec:intro}

Forward-looking sonar (FLS) forms an image by emitting an acoustic pulse into the environment and recording the returned backscatter over angle and time~\cite{Belcher1999DIDSON,Leonard2016AUVNavigation}. It is often the only reliable sensor in underwater settings, with uses including robotics and ecological monitoring. Yet, its images collapse elevation information, producing severe ambiguities, shadows, and noise. For recovering bathymetry, FLS offers advantages for local, short-range environments, \eg mapping a narrow inlet or a confined river segment. 

Multi-view pipelines and sensor fusion can eventually recover elevation from sonar~\cite{Negahdaripour2020FLSGeometry}, but they often require costly deployments, repeated passes, or overlapping sensor coverage. In many field scenarios, including long-term ecological monitoring, practitioners instead rely on a single fixed sonar viewpoint. Being able to infer geometry from a single view enables rapid on-site interpretation, continuous remote observation, and lower system complexity.

Perception-as-generation treats inference as fitting a scene model by simulating candidate states and comparing them to raw sensor measurements. In our setting, the scene is an explicit seafloor height field aligned with the sensor's polar sampling grid. We treat the unknown bathymetry as a latent representation that we iteratively render and refine until the synthetic rendering matches the observed sensor reading. As we explicitly simulate the forward model and parameterize it by real sonar settings, we can adapt to different sensor setups (\eg ranges and beam patterns) while remaining training-free. This gives our system the ability to transfer  readily between settings.
\begin{figure*}[t!]
  \centering
  \includegraphics[width=0.95\linewidth]{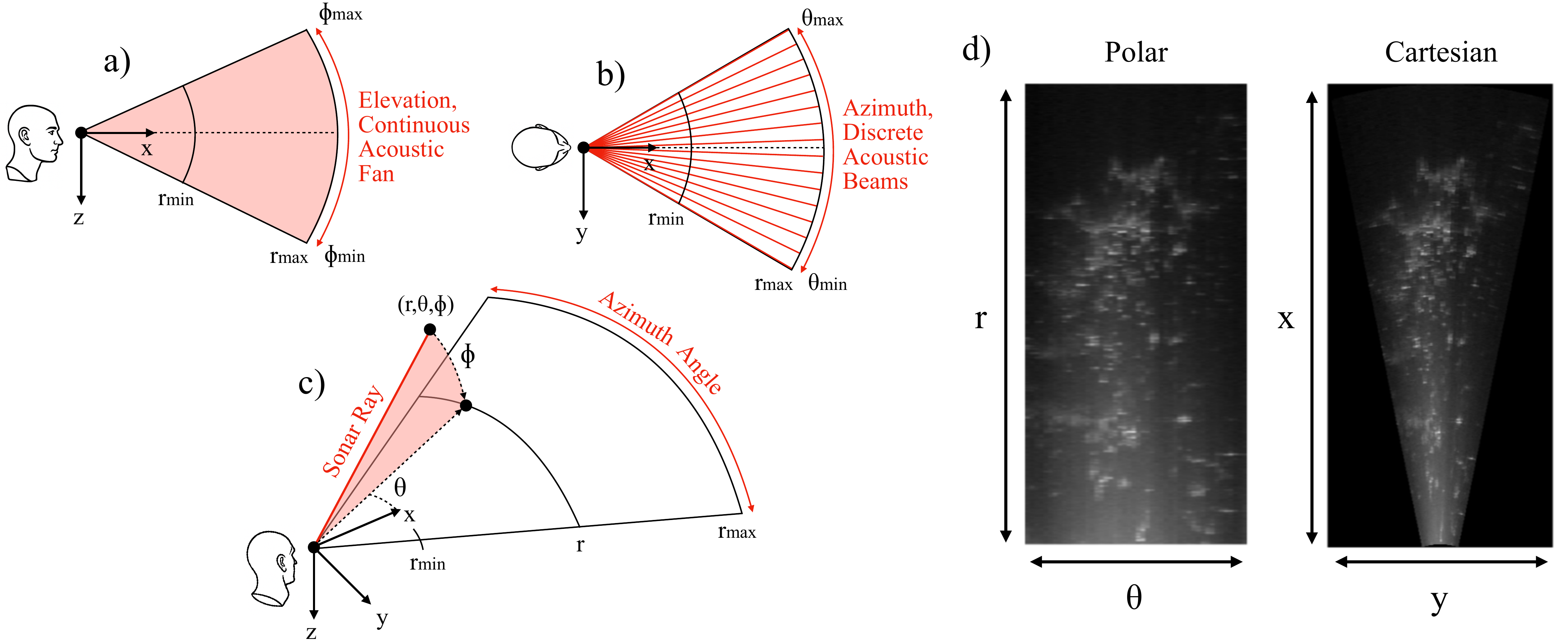}
  \caption{{\bf Overview of the sonar geometric layout from (a) a side view, (b) a top-down view, and (c) an isometric point of view}. As described in Sec.~\ref{sec:imaging_sonar_bg}, the forward-looking sonar (FLS) sampling process can be thought of as a discrete set of azimuthal beams, each comprised of a continuous acoustic signal in the elevation dimension. The signal returns are projected onto the horizontal plane, producing a top-down view of the scene with elevation ambiguity. (d) The sonar image can be viewed in polar coordinates, or warped into Cartesian. In the sonar image, pixel brightness encodes return intensity, while range is determined by the two-way travel time of the acoustic signal.}
  \label{fig:sonar_overview}
  \vspace{-10px}
\end{figure*}

Single-view bathymetry is well-known to be an ill-posed problem since many different scene geometries can explain the same sensor observation. Learning-based approaches, like CNNs, can form priors over a target distribution that allow them to estimate ambiguous geometry, but these priors are not usually invariant to changes in environment or sensor settings. We restrict our focus to settings where the underlying geometry is a height field and ground our reconstruction in principles of sonar physics that translate across domains. {Given the inherent ambiguity of the seafloor tilt in a single FLS view, we focus on recovering geometry conditioned on a known base-plane tilt.}

\vspace{4pt}
\noindent\textbf{Contributions.}
\begin{itemize}[leftmargin=*,nosep]
\item To our knowledge, the first application of differentiable rendering for height recovery in single-view sonar. 
\item A fast, training-free inverse method that optimizes a height field to reconstruct sonar images in \(<\)20 seconds.
\item A method that transfers across sonar configurations and environmental statistics by simulating the physical process with an explicit sensor parameterization. 
\end{itemize}

\section{Background and Related Work}
\label{sec:background}

The images generated by FLS can look deceptively similar to an optical camera, but the information that can be gained is dramatically different. The timing of returned signals can be converted into ranges, and readings are gathered across a spread of angles along a horizontal arc (azimuth). However, intensities are integrated over a vertical beam fan, meaning the elevation structure of the scene is collapsed into a flat image. Analogously to optical cameras, which provide width and height but not depth, a sonar image provides range and azimuth information but not altitudes.

\subsection{Imaging Sonar}
\label{sec:imaging_sonar_bg}
The acoustic pulse in FLS can be viewed as a discrete set of azimuthal beams, each composed of a continuous vertical fan of rays. \cref{fig:sonar_overview} visualizes this process, showing the polar representation of FLS. The sonar camera is considered to be at the origin, looking down the -X axis. The observed pixel intensities are determined by first–surface intersections along each beam, modulated by range–dependent transmission loss, noise, and multi-path and incidence effects. With a known speed of sound, the time-of-flight converts each return to range, reducing meter-scale ambiguity and providing pixel distances.

Sonar returns are typically dominated by diffuse and specular reflections, with relatively low ambient contributions. In natural settings, most surfaces can be well approximated using only diffuse reflections, while objects such as ship hulls and smooth rocks require specular consideration. In scenes with large, smooth, specular boundaries (\eg indoor tanks), specular reflections can produce multi-path returns, though this is less common in natural environments~\citep{Wang2023Multipath}. In practice, the signal is more likely to scatter, producing speckle. As such, compared to optics, sonar images typically have significant speckle noise (from scattering and interference) and low signal-to-noise ratio~\cite{Huang2020}.

\subsection{Sonar 3D Recovery}
Before modern learning-based pipelines, early work in sonar aimed to recover 3D shape directly from intensity using methods such as \textit{space carving} and \textit{shape from shading} (SfS). Space carving begins from a volumetric enclosure and iteratively removes voxels that fail multi-view consistency tests, leaving a volume that agrees with all views~\cite{Kutulakos1999SC}. Aykin and Negahdaripour~\cite{Aykin2015SC1, Aykin2017SC2} adapted this to sonar, but found it limited to simple shapes and heavily dependent on many viewpoints and rotations. SfS~\cite{Horn1970SfS} instead assumes a basic reflectance model, \eg Lambertian, and estimates surface normals or seafloor heights from brightness~\cite{Li1992SfS, Bikonis2013SfS1, Bikonis2013SfS2, Yakun2024SfS}. These approaches show that intensity encodes some 3D information, but only under restrictive assumptions and with extensive view coverage. Modern methods relax these assumptions by explicitly modeling the forward process or learning from data.

Sensor fusion, either using multiple sonar cameras or a combination of different sensor types, can also help resolve geometric ambiguities~\cite{Qadri2024AONeuS, Joe2019Fusion, Joe2022Fusion, Rho2025Fusion}, but it comes with increased hardware and calibration costs. In our work, we focus on 3D recovery from a single imaging sonar with a single view. DeBertoli \etal~\cite{DeBertoli2019ElevateNet} demonstrated that this is possible using a CNN trained to predict pixel-wise altitudes. Their method outperformed space carving, but required a large quantity of synthetic data, with results primarily demonstrated on simple objects.

\subsection{Differentiable Sonar Rendering}

Inverse rendering estimates scene parameters by matching rendered outputs to measurements. This optimization approach underlies differentiable rendering and NeRF-style methods for single- or multi-view reconstruction~\citep{Patow2003InvRendSurvey, Kato2020DiffRendSurvey, Mildenhall2020NeRF}. Recent work \citep{Qadri2022NeuSIS,Xie2022SSSDiff2, Bore2023SSSDiff, Reed2023SASDiff, Xie2022SSSDiff1,Xie2024BathymetryFLSDiff} has brought differentiable rendering to sonar. Qadri \etal~\citep{Qadri2022NeuSIS} demonstrated accurate 3D reconstruction of complex objects from multiple views using a volumetric rendering approach with neural signed distance functions (SDF), showing that physically-based forward models can achieve accurate shape recovery. However, their method depends on numerous views and is sensitive to complex scenes with multiple objects~\cite{Oliveira2025SES}. Additionally, representing the scene implicitly using neural networks increases the computational cost and fitting time. Comparatively, our method represents the seafloor as an explicit height field, allowing us to render a scene in under 100 milliseconds and fit a target image in tens of seconds. 

Forward models for elevation recovery have also been applied specifically for bathymetry, but are primarily limited to side-scan sonar (SSS) and synthetic aperture sonar (SAS)~\cite{Xie2022SSSDiff2, Bore2023SSSDiff, Reed2023SASDiff, Xie2022SSSDiff1}. Xie \etal~\cite{Xie2024BathymetryFLSDiff} introduced a self-supervised bathymetry framework that represents the seafloor as a neural heightmap (with multi-resolution hash encoding) and renders FLS intensities via a volumetric model that integrates along the elevation arc. Their method demonstrated accurate seafloor height recovery, but depends on many views along a surveying path. Unlike learned volumetric frameworks, we render images with an analytic, physics-based surface-scattering forward model.

\vspace{4pt}
\noindent\textbf{Our Approach.}
Existing methods for recovering 3D geometry from sonar typically rely on multiple views, target scenes with a single object, or require significant training data. Our focus is single–view FLS bathymetry using a physically grounded differentiable sonar simulator. Compared to other methods, we directly optimize a seafloor height field and fit observations using a differentiable physics approach, bypassing repeated neural-field evaluations during optimization while retaining adaptability across domains.
\section{Imaging-Sonar Forward Model}
\label{sec:forward-model}

Understanding how FLS forms an image is essential to reconstructing geometry from it. Our forward model treats the sonar readings as a collection of narrow azimuthal beams, each composed of a dense vertical fan of rays. The renderer explicitly models how these rays interact with the seafloor, how their returns combine into range–azimuth bins, and how physical factors such as incidence, gain, and spreading determine pixel intensities. 

\subsection{Seafloor Representation}
\label{sec:seafloor_rep}

We represent the seafloor as a discretized height field which can either be a Cartesian grid, or radially aligned with the sonar's azimuth beams. We place the sonar at the origin of a right-handed sensor frame, looking down -X with +Y right and +Z up (see Fig.~\ref{fig:sonar_overview}). All distances are in meters, and angles are in radians.

\begin{figure}[t]
  \centering
  \begin{minipage}{\columnwidth}
    \centering
    \includegraphics[width=1.0\columnwidth]{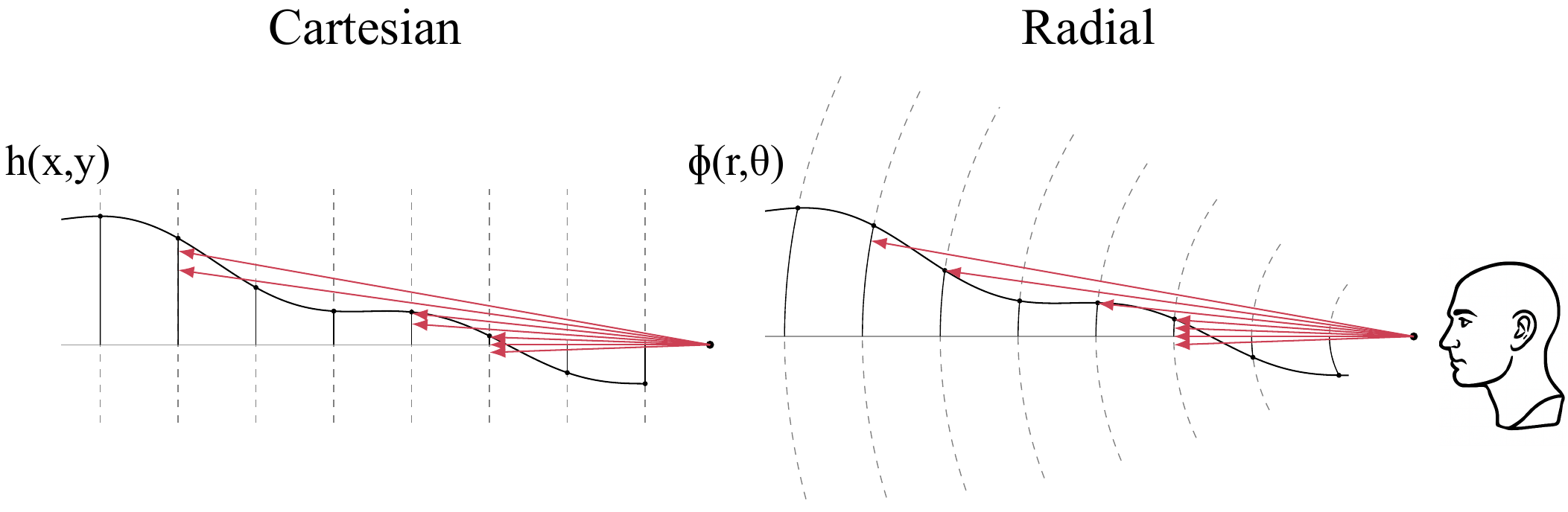}
    \caption{{\bf Side-view of the Cartesian (left) and polar (right) parameterizations} (\textit{cf.} Sec.~\ref{sec:seafloor_rep}). The polar height field stores angular heights, which vary along spherical shells of fixed radii. Ray intersections are determined by the first height entry which exceeds the height of the ray at that planar/spherical boundary.}
    \label{fig:heightfield_sideview}
  \end{minipage}
  
  \vspace{1em}
  
  \begin{minipage}{\columnwidth}
    \centering
    \includegraphics[width=1.0\columnwidth]{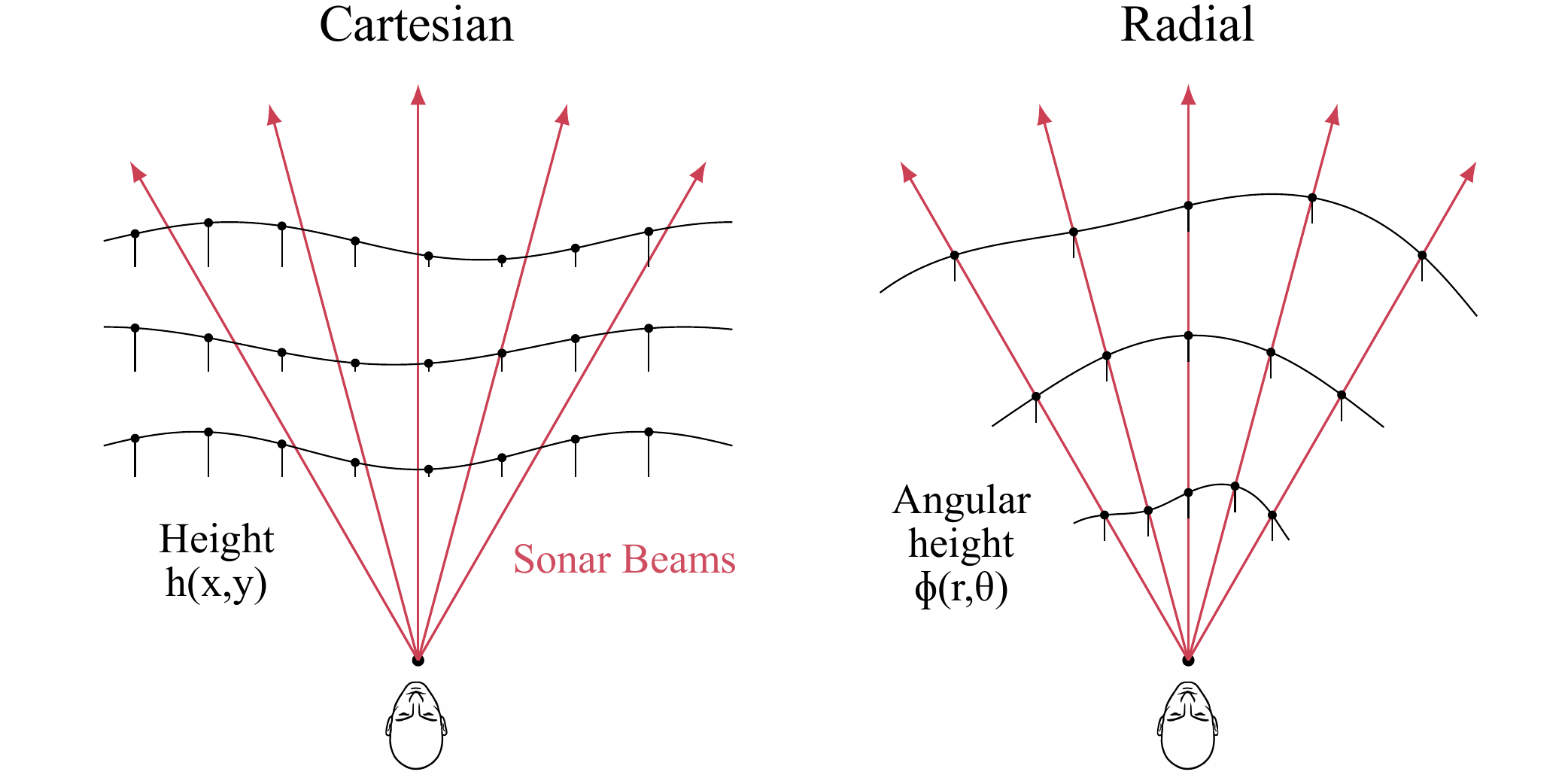}
    \caption{{\bf Top-view of the Cartesian (left) and polar (right) parameterizations} (\textit{cf.} Sec.~\ref{sec:seafloor_rep}). The polar field is aligned to the FLS sampling grid.\vspace{-8mm}}
    \label{fig:heightfield_topview}
  \end{minipage}
  
  \vspace{1.5em}
  
\end{figure}

As shown in Fig. \ref{fig:heightfield_sideview}, in the Cartesian representation each entry in the height field is a perpendicular displacement from the X-Y plane. When using the polar representation we instead use \textit{angular heights} measured in radians. Therefore, as the height at a given point is varied it remains on a spherical shell of constant radius.

In Fig. \ref{fig:heightfield_topview} we demonstrate how the height field is aligned with the sensor beams in the polar setting. Each column in the 2D height field corresponds to an azimuthal angle, which is chosen to match the angles sampled by the sonar sensor. This formulation has the advantage of being more stable in the learning process, since changes in heights remain local in their effect on pixel brightnesses.

\subsection{Intersections and Reflectance}
\label{sec:reflectance}
We construct a dense ray fan for each beam sampled by the sonar. Given $n_\text{az}$ azimuth beams and $n_\text{el}$ elevation samples (default $n_\text{el}=6n_\text{bins}$), we uniformly sample $n_\text{el}$ rays along the vertical arclength of the sensor for each of the $n_\text{az}$ beams, for a total of $n_\text{az}\,n_\text{el}$ ray samples. In this way the beams are sampled at discrete angles, while the elevation sampling approximates a continuous fan.

In Cartesian representations, ray intersections can be estimated by interpolating heights at various ranges and comparing against ray heights at those locations. In the polar seafloor, intersections are even simpler. Since the ray travels along only a single column of the beam-aligned height field, intersections are determined by the first entry whose angular height exceeds the angular elevation component of the ray (see Fig.~\ref{fig:heightfield_sideview}). For inverse rendering these operations are made differentiable, as explained in Sec.~\ref{sec:differentiable_intersections}.

\looseness=-1
We model per-ray backscatter as a sum of diffuse and specular terms at the ray–surface intersection.
Let $\mathbf{n}_{r,\theta}$ denote the unit surface normal of the height field at \((r,\theta)\) and $\boldsymbol{\omega}_{\theta,\phi}$ the unit direction from the surface back to the sensor. 
Co-located FLS transmitter and receiver lets us parameterize reflectance by this single vector.
The diffuse factor is calculated as
\vspace{-1.5mm}
\begin{equation}
\mu = \mathrm{max}\:\!\big(0, \mathbf{n}_{r,\theta}\!\cdot\!\boldsymbol{\omega}_{\theta,\phi}\big),
\end{equation}
and the specular term is modeled as a reflection lobe,
\begin{equation}
\exp\left(-\frac{1-\boldsymbol{\omega}_\text{refl}\!\cdot\!\boldsymbol{\omega}_{\theta,\phi}}{\sigma_{\text{spec}}^2}\right), 
\end{equation}
which takes maximum value when \(\boldsymbol{\omega}_{\theta,\phi}=\boldsymbol{\omega}_\text{refl}\), where \(\boldsymbol{\omega}_\text{refl}\) is the reflection direction given by
\begin{equation}
\boldsymbol{\omega}_\text{refl} = 2(\mathbf{n}\!\cdot\!\boldsymbol{\omega}_{\theta,\phi})\,\mathbf{n} - \boldsymbol{\omega}_{\theta,\phi}.
\end{equation}
The final reflectivity is then
\begin{equation}
f(r, \theta, \phi) = 
\Big[
\mu^{\gamma}
\;+\;
\exp\!\Big(-\frac{1-\boldsymbol{\omega}_\text{refl}\!\cdot\!\boldsymbol{\omega}_{\theta,\phi}}{\sigma_{spec}^2}\Big)
\Big]\,J,
\label{eq:reflectance}
\end{equation}
where $\gamma$ controls the diffuse falloff ($\gamma{=}1$ is Lambertian) and $\sigma_{spec}$ sets the specular lobe width. A geometric correction factor accounts for range spreading and foreshortening at range \(r\) as:
\vspace{-2mm}
\begin{equation}
J = \frac{r^2}{\max(\mu,\varepsilon)},
\end{equation}
with a small $\varepsilon$ for numerical stability.

\subsection{Image Formation and Processing}
From per-ray hit ranges (\(r\)) and reflectivities, intensities are spread into image bins. Gaussian binning weights are calculated as
\vspace{-1mm}
\[
\omega_k(r)\propto \exp\!\left[-\frac{1}{2}\left(\frac{u(r)-k}{\sigma_{\text{bins}}}\right)^{\!2}\right],
\]
where $u(r)$ is the fractional bin coordinate and $\sigma_{\text{bins}}$ controls spread. We apply a \(1/r^4\) correction factor to the reflectivities based on range \(r\) to account for two-way spherical geometric spreading, but implement a differentiable time-varying gain correction which optionally undoes this attenuation, as is commonly implemented in sonar sensor processing~\cite{Maclennan1986TVG}. Summed intensities are converted to log-amplitudes (dB) using $10\log_{10}(\cdot)$ compression.
\section{Inverse Reconstruction}
\label{sec:inverse-reconstruction}
Xie \etal~\cite{Xie2024BathymetryFLSDiff} introduce a rendering equation for sonar, with pixel intensity at \((r,\theta)\) modeled as
\begin{equation}
I(r, \theta) = \int_{\phi_{\text{min}}}^{\phi_{\text{max}}} \operatorname{\beta}(\theta, \phi) \operatorname{\sigma}(r, \theta, \phi) \operatorname{T}(r, \theta, \phi)  \operatorname{L}(r, \theta, \phi, \boldsymbol{v}) d \phi
\end{equation}
where \(\beta\) is the horizontal/vertical beam pattern, \(\sigma\) is the particle volume density, \(T\) is the accumulated transmittance along the ray path, and \(L\) is a learned radiance term evaluated using a neural network. 

In our setting (single-view FLS), we replace the volumetric radiance \(L\) with our surface backscatter model \(f\) (diffuse + specular) from Eq.~\ref{eq:reflectance}, removing repeated neural field evaluations during rendering. We further assume a separable beam pattern \(\beta(\theta,\phi)=\beta_\Theta(\theta)\beta_\Phi(\phi)\) and treat \(\beta_\Phi\) as uniform, yielding
\begin{equation}
I(r, \theta) = \beta_\Theta(\theta) \int_{\phi_{\text{min}}}^{\phi_{\text{max}}}  \operatorname{\sigma}(r, \theta, \phi) \operatorname{T}(r, \theta, \phi)  {f}(r, \theta, \phi) d \phi.
\end{equation}

\subsection{Differentiable Ray Intersections}
\label{sec:differentiable_intersections}
\looseness=-1
Focusing on the polar seafloor representation, we implement differentiable soft-intersections using logits $\phi_{\text{r},\theta}-\phi_\text{ray}$, where \(\phi_{\text{r},\theta}\) is the angular height of our latent seafloor at polar coordinate \((r, \theta)\) and \(\phi_\text{ray}\) is the angular elevation component of a given ray. For each ray, we calculate the logits for every entry down the corresponding column (azimuth beam) of our height field. The collision densities are calculated as
\begin{equation}
\sigma(r,\theta,\phi_\text{ray})=\sigmoid\!\big(\alpha(\phi_{\text{r},\theta}-\phi_{\text{ray}})\big),
\end{equation}
where \(\alpha\) is a scaling term which controls the sharpness of collision edges. 

With our range broken into discrete bins \(r_k\), the total probability of a ray transmitting to location \(j\) along its path is calculated as the cumulative probability of missing all previous entries down the beam:
\begin{equation}
    \operatorname{T}(r_j, \theta, \phi_\text{ray}) = \prod_{i<j}\!\Bigl(1-\sigma(r_i,\theta,\phi_\text{ray})\Bigr).
\end{equation}

\subsection{Beam-Pattern Modeling}
\label{sec:beam_pattern}
Beam-pattern effects are parameterized explicitly to model angle-dependent variations not explained by geometry. We find that learning a separate scaling parameter for each beam allows the model to more readily handle flicker effects or offscreen occlusions which can be averaged out in multi-view settings but become problematic for single-view reconstructions. To prevent interference with geometry fitting, we freeze these scalars for an initial warm-up period during optimization.

\subsection{Regularization and Priors}
Due to our Gaussian binning, high-frequency oscillations in the latent geometry can look equivalent to smooth regions. To discourage these artifacts we impose a first-order total-variation prior on the seafloor angular heights (\(\psi\)),
\begin{equation}
\mathrm{TV}(\psi)=\frac{1}{\lvert \psi\rvert}\sum_{i,j}\bigl\lvert \psi_{i+1,j}-\psi_{i,j}\bigr\rvert+\bigl\lvert \psi_{i,j+1}-\psi_{i,j}\bigr\rvert.
\end{equation}

In a single FLS frame, the tilt of the seafloor plane relative to the sensor is inherently ambiguous, since many orientations produce the same observed range support. Specifically, any plane that intersects the sensor arcs at \(r_{\min}\) and \(r_{\max}\) is a feasible orientation of the seafloor. However, the geometry is still locally constrained, since feasible base planes fall in a confined range of orientations and share similar sensor readings. Explanations that only fit the image at a knife-edge alignment from a specific view (\eg, explaining a shadow by just-parallel grazing) are considered non-generic, since other feasible tilts would be unlikely to produce the sensor reading.

Applying the generic viewpoint assumption~\cite{Freeman1994GVA}, we assume {that the rendering from our recovered geometry should remain stable under small perturbations of the base-plane tilt}. This discourages fragile, special-alignment interpretations and favors robust ones (\eg, interpreting stable shadows as depressions). Since this ambiguity is inherent in single-view sonar, we resolve geometry conditioned on known tilts, and use the generic-viewpoint prior to discourage interpretations that only explain the image under a single viewing angle.

\subsection{Loss and Optimization}
Given a processed target image, $I$, for a latent height field, \(\psi\), we simulate our differentiable ray logic, apply the learned azimuth gains, and push the result through our differentiable processing unit to get the final prediction $\hat{I}(\psi)$. We calculate the reconstruction loss as the per-pixel mean-squared error:
\vspace{-2mm}
\begin{equation}
\mathcal{L}_{\text{recon}}(\psi)=\frac{1}{HW}\left\lVert \hat{I}(\psi)-I\right\rVert_2^2.
\end{equation}
Under the generic viewpoint assumption, our optimization scheme becomes
\begin{equation}
\label{eq:main_loss}
    \min_{\psi} \mathbb{E}_{\phi_{\text{plane}}\sim p(\Phi)}[\mathcal{L}_{\text{recon}}(\psi)+\lambda_{\text{TV}}\operatorname{TV}(\psi)],
\end{equation}
where \(\Phi\) denotes the set of feasible base-plane orientations. In practice, we resample \(\phi_{\text{plane}}\) at each optimization step from a uniform distribution over planes which span at least 60\% of the elevation fan (typically representing only a few degrees of tilt range) and present a full ablation in the supplemental materials.

\begin{figure*}[t]
  \centering
   \includegraphics[width=1.0\linewidth]{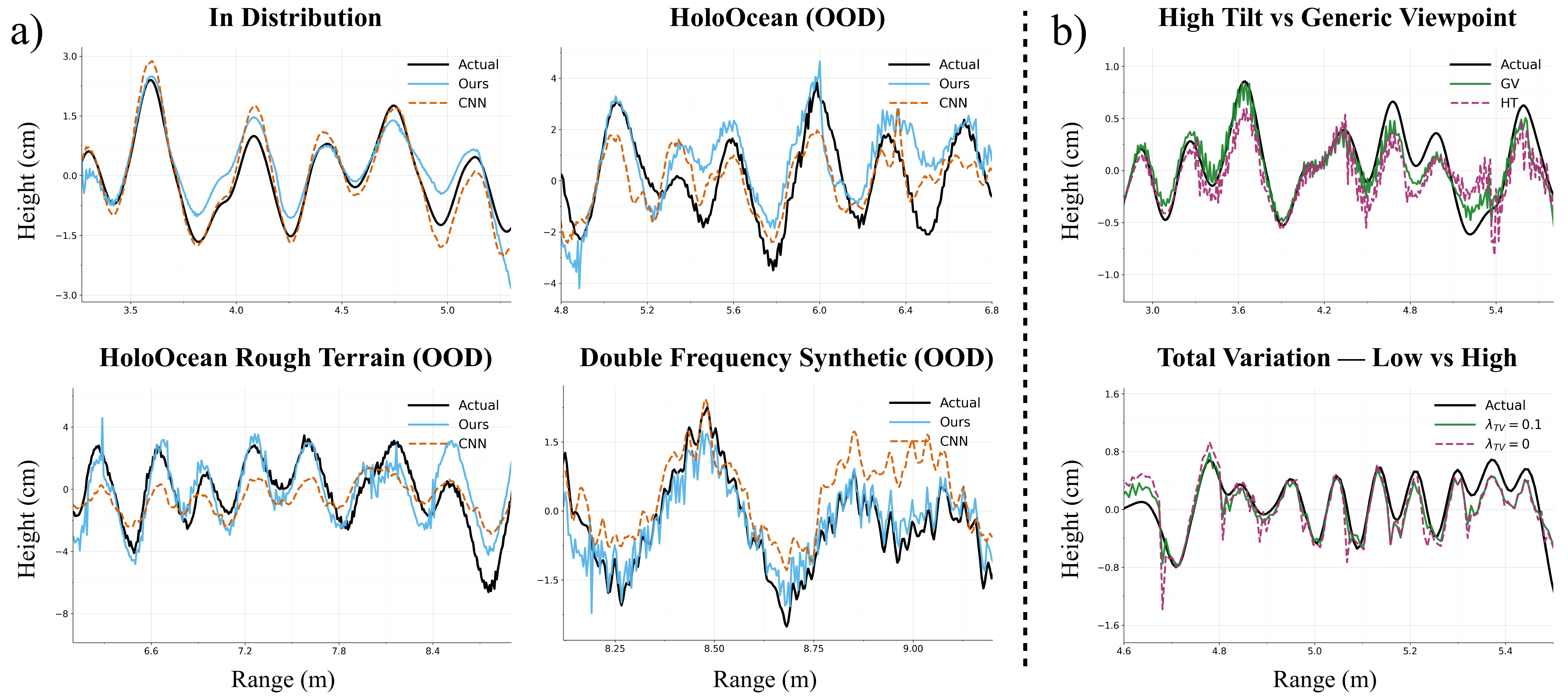}
\vspace{-0.2in}
   \caption{{\bf Heights predicted down a single azimuth beam as compared to the ground truth}. (a) Samples from each of our datasets, as well as a double-frequency case where we stack Perlin noise. (b) Comparisons between our generic viewpoint implementation and a fixed high-tilt plane, and low vs high TV regularization strengths. See Sec.~\ref{sec:qualitative} for details. \vspace{-4mm}}
   \label{fig:azimuth_preds}
\end{figure*}

\section{Experiments}
\label{sec:experiments}

We evaluate performance on synthetic datasets with known ground-truth elevations and compare against a supervised CNN baseline. {All methods are evaluated in a common 3D reference frame by converting predicted height deviations into point clouds defined on the ground-truth sonar sampling grid. We then report mean Chamfer distance (MCD) and 3D RMSE/MAE/MSE between point sets (\cref{sec:model_eval}).} Synthetic datasets are generated using a combination of our simulator and the HoloOcean sonar simulator built in Unreal Engine~\cite{2022PotokarHoloOcean,Romrell2025HoloOcean2}. We additionally test qualitative fits on real river imagery without ground-truth elevations.
 
\subsection{Synthetic In-Distribution Dataset}
\label{sec:in_dist_exp}
Using our simulator, we generated 10,000 training samples and 200 validation samples for training the CNN, and 200 testing samples to evaluate all methods. The training set was not used for our approach since we fit geometry on individual frames without pretraining. Seafloors were generated using Perlin noise with randomized frequency and centimeter-scale amplitudes added to a base-plane at randomized tilts. Each plane was chosen to ensure returns along the full sonar range. Each sample also had a randomized sonar configuration, including range, binning, azimuth coverage, and elevation angles. Ranges in consideration were \(<\)10m, and elevation coverage was chosen in the range of \(10-30^{\circ}\). Full dataset parameters are listed in the supplemental materials. Each validation sample input is comprised of the sensor reading and sonar configuration metadata. The target is a height-deviation map defined on the sample's sonar range–azimuth grid in the coordinate frame of the sample's base plane (\ie, conditioned on the known base-plane tilt).

\subsection{HoloOcean Datasets}
To evaluate out-of-distribution performance, we generated two additional test datasets using HoloOcean, each with samples in the same input-output format as in Sec.~\ref{sec:in_dist_exp}. In the first dataset we use a similar randomized Perlin noise and sonar configuration sampling scheme as the in-distribution data. We also generated a ``rough terrain'' dataset with increased amplitudes and range coverage, making the task more challenging. 

\subsection{Real River Fits}
Targets from real riverbeds were acquired using ARIS and Didson sonar cameras, with samples produced by averaging the frames over a clip to reduce noise and transitory objects. The corresponding configuration files were parsed to initialize our model with the proper beam directions, samples per beam, and range parameters. No processing is done to the raw target frames other than rescaling. Images were collected from a set of rivers near the U.S. West Coast.

\subsection{CNN Baseline}
For the CNN baseline, we use a standard U-Net with four encoder and four decoder stages (3$\times$3 conv--BN--ReLU blocks), a 1/16-resolution bottleneck with 1024 channels, and bilinear upsampling with skip connections to predict a single-channel height-deviation map. The model is trained for 200 epochs on 10{,}000 samples using MSE loss.

\subsection{Model Evaluation}
\label{sec:model_eval}

We evaluate models from their predicted height-deviation map on the range--azimuth grid by converting predictions and ground truth into 3D point clouds in a common coordinate frame. For each bin, we compute its 3D location on the base plane using the known tilt and sensor geometry, then offset along the local radial direction in the sensor's range–azimuth plane (\ie, along the constant-range arc at that bin).

\subsection{Implementation Details}
We implement our simulator using PyTorch~\cite{Paszke2019Pytorch} for autodiff., and we fit our seafloor height field by optimizing Eq.~\ref{eq:main_loss} using AdamW~\cite{2019LoshchilovAdamW} for 150 steps using a learning rate of \(1e-4\). Azimuth scaling factors have a learning rate of \(0\) until \(n_{\text{warmup}}=30\) steps when we raise it to \(0.1\). We set \(\lambda_{\text{TV}}=0.1\) by default. On an RTX 5090 our model achieves ~12 iterations per second, generating an image in under \(100\)ms. Fully fitting a 3D height field takes \(5-15\)s for \(150\) optimization steps, depending on target size.

In all synthetic experiments, the base-plane tilt is treated as known at evaluation time, and the CNN is trained and evaluated under the same assumption. We report three variants of our inverse method that differ only in the {optimization-time} plane orientation: known-plane (KP) uses the known tilt, high-tilt (HT) fixes a steep tilt for all samples, and generic-view (GV) resamples a new tilt each optimization step to encourage robust geometry.
\begin{table}[t]
  \centering
  \caption{
    \centering
        HoloOcean Results.\\
      HT = High Tilt, KP = Known Plane, GV = Generic Views.\\
      MCD, RMSE, \& MAE are in cm, and MSE is cm$^2$.\\
      CNN results are averaged over five runs.}
  \label{tab:holoocean}
  \begin{tabular}{lcccc}
    \toprule
    Method & MCD ↓ & RMSE ↓ & MAE ↓ & MSE ↓ \\
    \midrule
    CNN & 0.802 & 1.066 & 0.851 & 1.212 \\
    Ours (HT) & 0.694 & 0.928 & 0.750 & 0.969 \\
    Ours (KP) & 0.690 & 0.921 & 0.745 & 0.956 \\
    Ours (GV) & \textbf{0.671} & \textbf{0.830} & \textbf{0.717} & \textbf{0.900} \\
    \bottomrule
  \end{tabular}
\end{table}
\begin{table}[t]
  \centering
  \caption{HoloOcean Rough Terrain Results.}
  \label{tab:holoocean_rough}
  \begin{tabular}{lcccc}
    \toprule
    Method & MCD ↓ & RMSE ↓ & MAE ↓ & MSE ↓ \\
    \midrule
    CNN & 1.340 & \textbf{1.988} & \textbf{1.606} & \textbf{4.075} \\
    Ours (HT) & 1.363  & 2.099 & 1.685 & 4.499 \\
    Ours (KP) & 1.347 & 2.067 & 1.662 & 4.358 \\
    Ours (GV) & \textbf{1.333} & 2.032 & 1.636 & 4.220 \\
    \bottomrule
  \end{tabular}
\end{table}
\begin{table}[t]
  \centering
  \caption{
    \centering
      In-Distribution Synthetic Results.}
  \label{tab:in_dist}
  \begin{tabular}{lcccc}
    \toprule
    Method & MCD ↓ & RMSE ↓ & MAE ↓ & MSE ↓ \\
    \midrule
    CNN & \textbf{0.330} & \textbf{0.439} & \textbf{0.337} & \textbf{0.229} \\
    Ours (HT) & 0.504 & 0.677 & 0.528 & 0.586 \\
    Ours (KP) & 0.517 & 0.692 & 0.541  & 0.598 \\
    Ours (GV) & 0.504 & 0.674 & 0.527 & 0.577 \\
    \bottomrule
  \end{tabular}
  \vspace{-3mm}
\end{table}

\section{Results and Discussion}
\label{sec:results}

Our model achieves {competitive} performance without any training, {particularly under distribution shift}, and can fit sonar images across varying real-world setups. We evaluate our method quantitatively as well as qualitatively by plotting heights across single azimuth beams.

\subsection{Quantitative Results}
\label{sec:quantitative}
On the in-distribution synthetic dataset, the CNN achieves lower error than our method (Table~\ref{tab:in_dist}), reflecting the strength of supervised priors when training and test conditions align. In this setting, a network trained on abundant labeled data can leverage domain-specific statistics that our single-view inverse method does not model.

Out of distribution, the CNN’s advantage shrinks considerably. On the standard HoloOcean benchmark, our method outperforms the CNN across all metrics (Table~\ref{tab:holoocean}), with GV best and the others close behind. On rough terrain, the results are essentially tied. We match or beat on MCD, while the CNN is slightly better on RMSE/MAE/MSE, with ours close behind (Table~\ref{tab:holoocean_rough}). Overall, explicitly modeling the sonar forward process improves robustness under changes in setup and seafloor statistics, while learned priors remain strongest in-distribution.
    
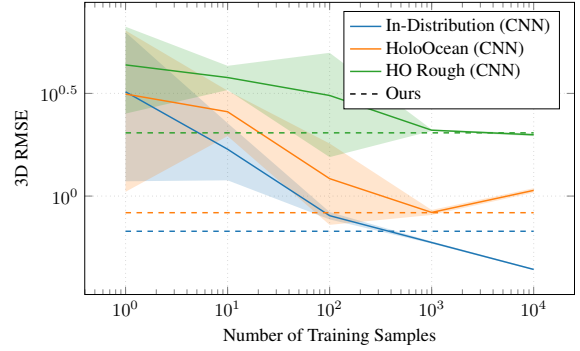
\begin{figure}
  \centering
  \resizebox{1.0\columnwidth}{!}{
    \definecolor{ds1}{HTML}{1F77B4}
\definecolor{ds2}{HTML}{FF7F0E}
\definecolor{ds3}{HTML}{2CA02C}
\definecolor{ds4}{HTML}{D62728}
\definecolor{ds5}{HTML}{9467BD}
\definecolor{ds6}{HTML}{8C564B}
\definecolor{ds7}{HTML}{E377C2}
\definecolor{ds8}{HTML}{7F7F7F}
\definecolor{ds9}{HTML}{BCBD22}
\definecolor{ds10}{HTML}{17BECF}
\definecolor{blk}{HTML}{000000}
\begin{tikzpicture}
\begin{axis}[
  width=11cm, height=7.2cm,
  xmode=log, ymode=log,
  xmajorgrids, ymajorgrids,
  xminorgrids=false, yminorgrids=false,
  grid style={line width=0.3pt, dotted},
  xlabel={Number of Training Samples},
  ylabel={3D RMSE},
  title={},
  legend pos=north east,
  legend cell align=left,
]

\addplot[name path=ds1upper, draw=none, forget plot] table[row sep=\\] {%
x y \\
1.0 6.2761678875137505 \\
10.0 2.264764485778823 \\
100.0 0.8363415405967946 \\
1000.0 0.601405598162851 \\
10000.0 0.442883352155721 \\
};
\addplot[name path=ds1lower, draw=none, forget plot] table[row sep=\\] {%
x y \\
1.0 1.1804740528309197 \\
10.0 1.1931449773582334 \\
100.0 0.7700387148967665 \\
1000.0 0.5868172736746164 \\
10000.0 0.43573466698905 \\
};
\addplot[fill=ds1, fill opacity=0.2, forget plot] fill between[of=ds1upper and ds1lower];
\addplot[{ds1}, thick] table[row sep=\\] {%
x y \\
1.0 3.218517 \\
10.0 1.6909732000000002 \\
100.0 0.802862 \\
1000.0 0.5940898 \\
10000.0 0.43930199999999997 \\
};
\addlegendentry{In-Distribution (CNN)}
\addplot[name path=ds2upper, draw=none, forget plot] table[row sep=\\] {%
x y \\
1.0 6.402212704048201 \\
10.0 3.2607351124639457 \\
100.0 1.8031339278206613 \\
1000.0 0.8559764199750425 \\
10000.0 1.0920224716340774 \\
};
\addplot[name path=ds2lower, draw=none, forget plot] table[row sep=\\] {%
x y \\
1.0 1.0490848632319167 \\
10.0 1.9591812386303986 \\
100.0 0.7240756453294922 \\
1000.0 0.8092475853024645 \\
10000.0 1.0403611117407856 \\
};
\addplot[fill=ds2, fill opacity=0.2, forget plot] fill between[of=ds2upper and ds2lower];
\addplot[{ds2}, thick] table[row sep=\\] {%
x y \\
1.0 3.1393694 \\
10.0 2.5740644 \\
100.0 1.2158598 \\
1000.0 0.8324556 \\
10000.0 1.0660409999999998 \\
};
\addlegendentry{HoloOcean (CNN)}
\addplot[name path=ds3upper, draw=none, forget plot] table[row sep=\\] {%
x y \\
1.0 6.6800615396591185 \\
10.0 4.307812496303215 \\
100.0 4.984460158973107 \\
1000.0 2.111594988564347 \\
10000.0 2.019017922366926 \\
};
\addplot[name path=ds3lower, draw=none, forget plot] table[row sep=\\] {%
x y \\
1.0 2.5213891513397475 \\
10.0 3.282740788284345 \\
100.0 1.5476577707962822 \\
1000.0 2.075171222935963 \\
10000.0 1.9576098152701882 \\
};
\addplot[fill=ds3, fill opacity=0.2, forget plot] fill between[of=ds3upper and ds3lower];
\addplot[{ds3}, thick] table[row sep=\\] {%
x y \\
1.0 4.3533072 \\
10.0 3.778242 \\
100.0 3.0896438 \\
1000.0 2.0933450000000002 \\
10000.0 1.9881997999999999 \\
};
\addlegendentry{HO Rough (CNN)}
\addlegendimage{black, thick, dashed}
\addlegendentry{Ours}
\addplot[{ds1}, thick, dashed] coordinates {(1.0,0.674) (10000.0,0.674)};
\addplot[{ds2}, thick, dashed] coordinates {(1.0,0.830) (10000.0,0.830)};
\addplot[{ds3}, thick, dashed] coordinates {(1.0,2.032) (10000.0,2.032)};

\end{axis}
\end{tikzpicture}
  }
  \vspace{-0.2in}
  \caption{{\bf CNN accuracy on the three test sets vs. training set size.} Solid lines show the mean over 5 independent training runs per sample size, and shaded regions show the 95\% CI. Our training-free method is shown as a constant dashed line.\vspace{-0.15in}}
  \label{fig:cnn_logplot}
\end{figure}

To examine the tradeoff between supervision and forward modeling, we vary the CNN training set size in \cref{fig:cnn_logplot}. CNN performance generally degrades as labels become scarce, while our method is unchanged. In distribution the CNN approaches our error between \(10^2\) and \(10^3\) samples, and at smaller dataset sizes our method is consistently better. CNN training uses only in-distribution train/val splits with early stopping (no OOD access).

\subsection{Qualitative Results}
\label{sec:qualitative}
Our model recovers the overall seafloor structure across domains without any training. We visualize predicted heights down a single azimuth beam for several reconstructions across our datasets in Fig.~\ref{fig:azimuth_preds}. {Our method regularly produces heights close to the ground truth, and generally demonstrates greater adaptability than the CNN when we stack frequencies to generate more complex geometry.}

\begin{figure}
  \centering
  \includegraphics[width=1.0\columnwidth]{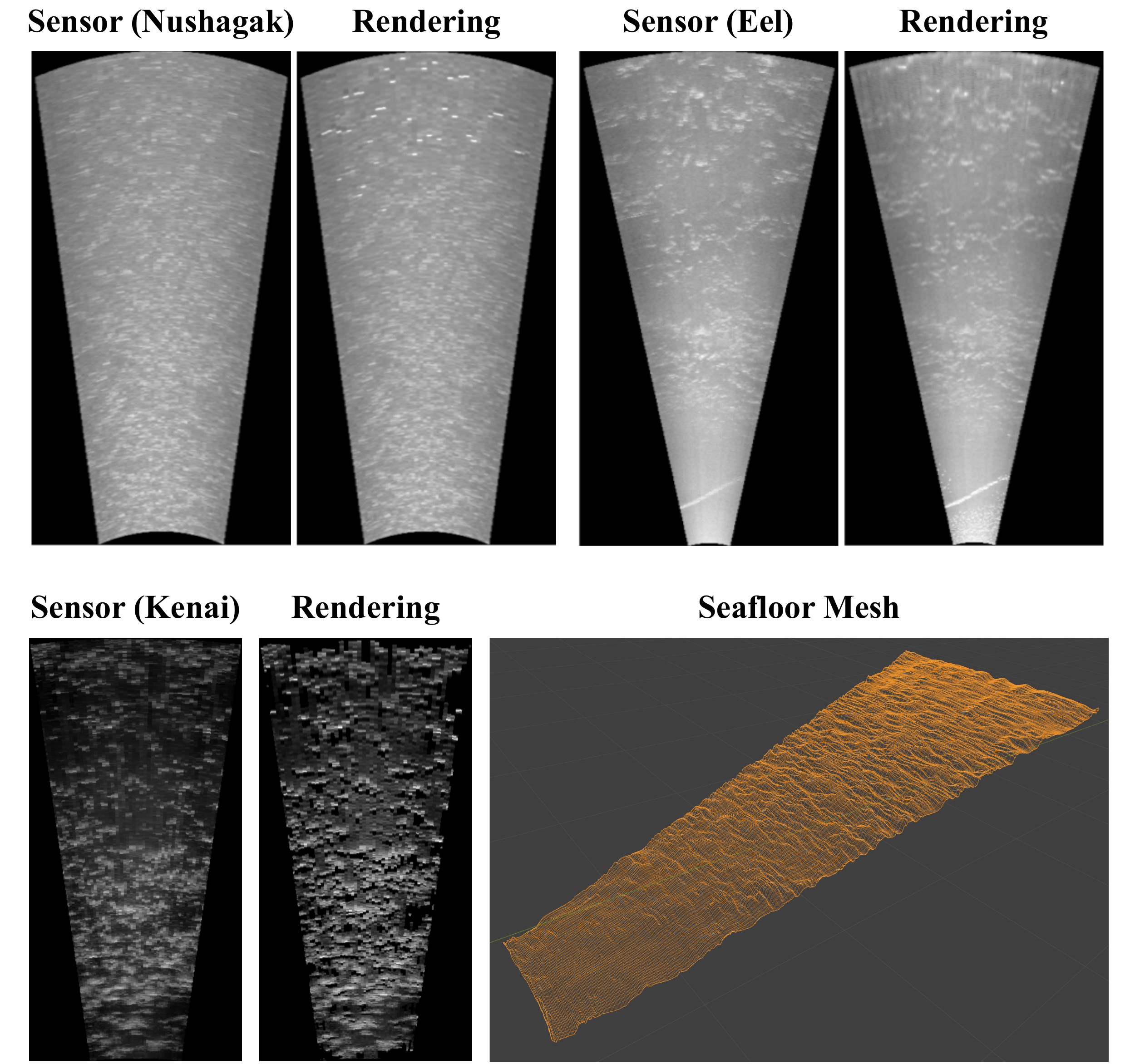}
  \caption{{\bf Sensor readings and reconstructed images using sonar images from real rivers} (\textit{cf.} Sec.~\ref{sec:qualitative}). From top-left to bottom are the Nushagak, Eel, and Kenai rivers. Additional 3D reconstructions are displayed in the supp. materials.}
  \label{fig:river_fits}
\end{figure}

To highlight the applicability of our approach, we demonstrate image reconstructions for sensor targets taken from real riverbeds across the U.S. West Coast in Fig.~\ref{fig:river_fits}. The model recovers detailed image structure along with low-frequency texture and occlusions, suggesting that the inferred height fields capture plausible 3D geometry.
\subsection{Failure Cases and Limitations}
\label{sec:limitations}

\begin{figure}
  \centering
  \includegraphics[width=0.85\columnwidth]{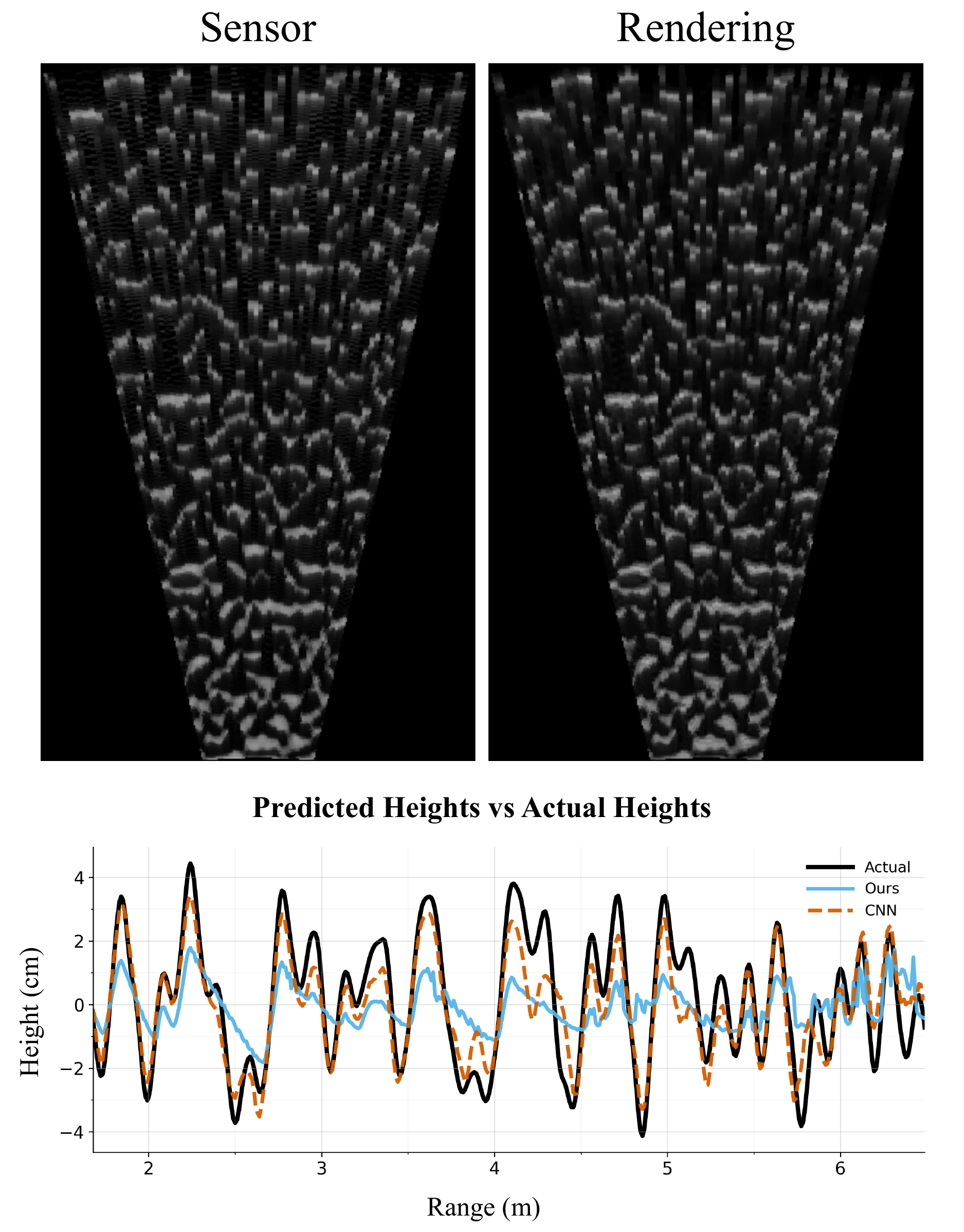}
  \vspace{-0.05in}
  \caption{{\bf Severe single-view ambiguities in imaging sonar}. Despite accurately reconstructing the sensor readings, this in-dist. sample's recovered heights are substantially shorter than the ground truth, as discussed in \cref{sec:limitations}. The CNN resolves this by learning the statistics of seafloor depth, and performs well.}
  \label{fig:fail_case}
  \vspace{-2mm}
\end{figure}

The primary limitation of our model lies in the inherent ambiguity present in single-view sonar imagery: many different geometries can explain the same observed sensor reading. In particular, our inverse renderer can achieve accurate recovery of geometric patterns, but tends to predict the minimum heights necessary to explain the observed data. Compared to a CNN, which can learn that occluded regions tend to have a certain depth, the height field gradients in our approach go to zero as the 2D image reconstruction approaches the target. This is both a strength and a limitation in our approach, since the differentiable renderer makes only the necessary geometric changes to explain observed data rather than relying on learned assumptions. We show this limitation explicitly in Fig.~\ref{fig:fail_case}, where we demonstrate a failure to recover the underlying heights even though the sensor reading is recovered accurately.

An additional limitation present in our differentiable renderer is that noise and artifacts in a target are explained by changes to the predicted geometry since we are optimizing for minimal reconstruction loss. This appears as high-frequency jumps in our predicted heights, which our TV regularization helps to reduce but doesn't fully eliminate.

\subsection{TV Ablation}

We perform an ablation over \(\lambda_\text{TV}\) under the generic-view (GV) setting (Table~\ref{tab:tv_ablation}). Our method remains competitive across small values, but the results clearly benefit from a moderate smoothness prior that suppresses high-frequency artifacts. For our tasks, the best performance occurs with \(\lambda_\text{TV}\!=\!0.01\) on the in-distribution dataset, and around \(1.0\) on the standard HoloOcean and rough-terrain benchmarks.

\begin{table}[t]
  \centering
  \caption{
    \centering
    Increasing the TV regularization increases the accuracy of our method on HoloOcean. Evaluated using the generic viewpoint method and 3D MSE (cm, ↓).}
  \label{tab:tv_ablation}
  \begin{tabular}{lcccc}
    \toprule
    \(\lambda_\text{TV}\) & In Dist. & HoloOcean & HO Rough \\
    \midrule
    0 & {0.577} & 0.930 & 4.246 \\
    1e-2 & \textbf{0.575} & 0.923 & 4.244 \\
    1e-1 & {0.578} & {0.902} & {4.200} \\
    1.0 & 0.633 & \textbf{0.880} & \textbf{4.170} \\
    1e1 & 0.913 & 1.172 & 5.175 \\
    1e2 & 1.280 & 1.657 & 7.245 \\
    \bottomrule
  \end{tabular}
  \vspace{-3mm}
\end{table}

\section{Conclusion}
\label{sec:conclusion}
\looseness=-1
We introduce a training-free method for single-view bathymetry in FLS that fits an explicit height field via differentiable physics-based rendering. By grounding inference in the forward sonar process and simple priors, our approach transfers across sensor configurations without training. On synthetic benchmarks, it outperforms or remains near-equal to a supervised CNN out of distribution, while the CNN remains best in distribution. On real rivers it reconstructs image structure across sites accurately. These results support inverse rendering as a path to single-view geometry recovery from imaging sonar and a foundation for systems that must interpret new environments quickly from sparse observations.

{
    \small
    \bibliographystyle{ieeenat_fullname}
    \bibliography{main}
}

\clearpage
\setcounter{page}{1}
\maketitlesupplementary

\section{Synthetic Dataset Parameters}
\label{sec:dataset_params}

\begin{figure*}[t!]
  \centering
  \includegraphics[width=0.75\linewidth]{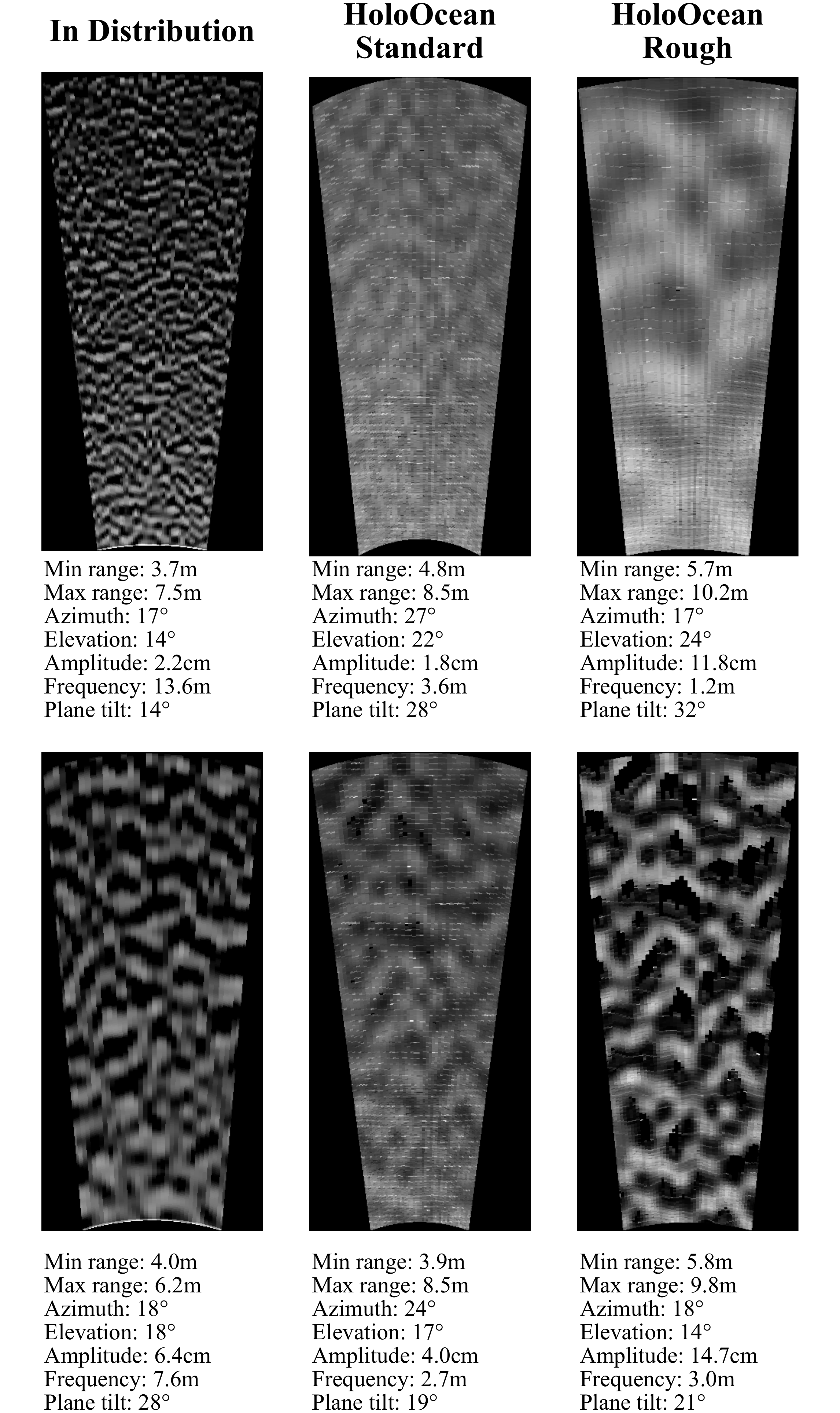}
  \caption{{\bf Example target images from each of our three datasets}. Sampling details can be found in Sec.~\ref{sec:dataset_params}}
  \label{fig:dset_samples}
\end{figure*}

\cref{tab:in_dist_sampling,tab:holoocean_standard_sampling,tab:holoocean_rough_sampling} summarize the empirical sampling ranges observed in each dataset. We report the minimum and maximum values across the full split for each setting. Note that the amplitude refers to the peak-to-trough height (maximum minus minimum elevation). Although we targeted matching configuration ranges between the synthetic and HoloOcean standard datasets, integration with HoloOcean resulted in slightly different realized parameter bounds. We present samples from each dataset in Fig.~\ref{fig:dset_samples}.

\begin{table}[t]
  \centering
  \caption{\centering
    { In-distribution synthetic sampling parameters.}\\
    Empirical min/max over 10{,}000 train, 200 val, \\ and 200 test samples.}
  \label{tab:in_dist_sampling}
  \begin{tabular}{@{}lcc@{}}
    \toprule
    Parameter & Min & Max \\
    \midrule
    \multicolumn{3}{@{}l}{\textbf{Sensor parameters}} \\
    Azimuth spread (\(^\circ\))      & 10.0 & 19.0 \\
    Start range (m)                  & 1.53 & 4.97 \\
    End range (m)                    & 3.80 & 7.50 \\
    Range coverage (m)               & 2.03 & 5.49 \\
    Range bins                       & 380  & 512  \\
    Azimuth bins                     & 36   & 64   \\
    Elevation spread (\(^\circ\))    & 10.0 & 19.0 \\
    \midrule
    \multicolumn{3}{@{}l}{\textbf{Seafloor parameters}} \\
    Amplitude (cm)                   & 2.01 & 9.98 \\
    Frequency (\(\text{m}^{-1}\))   & 2.00 & 15.0 \\
    Ground tilt (\(^\circ\))         & 6.87 & 38.0 \\
    \bottomrule
  \end{tabular}
\end{table}

\begin{table}[t]
  \centering
  \caption{\centering
    { HoloOcean standard sampling parameters.}\\
    Empirical min/max over 149 samples.}
  \label{tab:holoocean_standard_sampling}
  \begin{tabular}{@{}lcc@{}}
    \toprule
    Parameter & Min & Max \\
    \midrule
    \multicolumn{3}{@{}l}{\textbf{Sensor parameters}} \\
    Azimuth spread (\(^\circ\))      & 10.0 & 29.0 \\
    Start range (m)                  & 1.01 & 5.99 \\
    End range (m)                    & 4.16 & 8.50 \\
    Range coverage (m)               & 2.51 & 5.50 \\
    Range bins                       & 512  & 512  \\
    Azimuth bins                     & 48   & 48   \\
    Elevation spread (\(^\circ\))    & 10.0 & 29.0 \\
    \midrule
    \multicolumn{3}{@{}l}{\textbf{Seafloor parameters}} \\
    Amplitude (cm)                   & 1.03 & 7.97 \\
    Frequency (\(\text{m}^{-1}\))   & 1.00 & 4.00 \\
    Ground tilt (\(^\circ\))         & 6.59 & 47.23 \\
    \bottomrule
  \end{tabular}
\end{table}

\begin{table}[t]
  \centering
  \caption{\centering
    { HoloOcean rough-terrain sampling parameters.}\\
    Empirical min/max over 35 samples.}
  \label{tab:holoocean_rough_sampling}
  \begin{tabular}{@{}lcc@{}}
    \toprule
    Parameter & Min & Max \\
    \midrule
    \multicolumn{3}{@{}l}{\textbf{Sensor parameters}} \\
    Azimuth spread (\(^\circ\))      & 10.0 & 29.0 \\
    Start range (m)                  & 5.14 & 7.98 \\
    End range (m)                    & 8.56 & 12.62 \\
    Range coverage (m)               & 3.12 & 4.97 \\
    Range bins                       & 512  & 512  \\
    Azimuth bins                     & 48   & 48   \\
    Elevation spread (\(^\circ\))    & 11.0 & 28.0 \\
    \midrule
    \multicolumn{3}{@{}l}{\textbf{Seafloor parameters}} \\
    Amplitude (cm)                   & 10.0 & 15.9 \\
    Frequency (\(\text{m}^{-1}\))   & 1.00 & 3.50 \\
    Ground tilt (\(^\circ\))         & 14.4 & 51.1 \\
    \bottomrule
  \end{tabular}
\end{table}

We refer to the synthetic dataset generated with our own forward model as ``in-distribution,'' since the CNN is trained and evaluated on samples drawn from the same simulator and parameter ranges. The HoloOcean datasets differ in simulator, sensor model, and realized parameter ranges, and are therefore treated as out of distribution, even though the underlying seafloor prior (Perlin height fields) is shared.

\section{Hyperparameters and Configs}
\label{sec:hyperparameters}

\begin{table*}[t]
    \centering
    \caption{\centering
        Key hyperparameters for differentiable rendering and inverse reconstruction.}
    \label{tab:hyperparams}
    \begin{tabular}{@{}lll@{}}
        \toprule
        Component & Parameter & Default / range \\
        \midrule
        Forward model / renderer
        & Elevation samples \(n_\mathrm{el}\) &
            \(6\,n_\mathrm{bins}\) per azimuth beam \\
        & Gaussian bin spread \(\sigma_\mathrm{bins}\) &
            \num{0.5} - \num{1.0} bins \\
        & Near-range padding &
            4 bins in front of \(r_\mathrm{min}\) \\
        & Two-way geometric spreading &
            \(1/r^4\) correction (TVG undo optional) \\
        & TVG exponent &
            \num{3.2} - \num{4.0} \\
        & Diffuse exponent \(\gamma\) &
            \num{1.0} - \num{2.0} \\
        & Specular spread \(\sigma_\mathrm{spec}\) &
            \(5^\circ - 10^\circ\) \\
        \midrule
        Seafloor / intersections
        & Collision sharpness \(\alpha\) &
            \num{2500} - \num{4500} \\
        & HT plane coverage &
            \(90\%\) of elevation span \\
        & GV sampling range &
            60–97.5\% of elevation span \\
        \midrule
        Image processing
        & Azimuth gains &
            One scalar per beam \\
        & Gain learning rate &
            \(0 \rightarrow 10^{-1}\) (after warmup) \\
        \midrule
        Optimization
        & Steps per frame &
            \num{100}-\num{300} \\
        & TV weight \(\lambda_\mathrm{TV}\) &
            \num{0.0} - \num{1.0} \\
        & Optimizer &
            AdamW \\
        & Geometry learning rate &
            \(10^{-4}\) \\
        & Warmup steps \(n_\mathrm{warmup}\) &
            \num{30} (freeze gains) \\
        \bottomrule
    \end{tabular}
\end{table*}

Table~\ref{tab:hyperparams} provides the key hyperparameter ranges for our differentiable renderer and inverse reconstructions. We choose the elevation sampling density, \(n_\text{el}\), so that each beam approximates a continuous vertical fan. The near-range padding adds out-of-view bins to the front of the height field to prevent artifacts from offscreen occlusions. The ranges listed here include values that we found stable in practice. We did not tune these heavily for each dataset.

\section{Additional Ablations}
\label{sec:more_ablations}
To further identify the optimal configurations of our system, and the benefits of the generic viewpoint approach, we perform additional ablations over the learning rate, optimization steps, and generic viewpoint coverage.

\subsection{Learning Rate Ablation}
\label{sec:lr_ablate}

To study the tradeoff between optimization budget and reconstruction accuracy we sweep the geometry learning rate across several orders of magnitude and vary the number of gradient steps. For each combination we fit the polar height field on the HoloOcean rough terrain dataset using the generic viewpoint (GV) approach and a fixed TV weight of \num{0.1}. We report the 3D MSE over the test set.

\begin{table*}[h]
  \centering
  \captionsetup{width=0.85\linewidth}
  \caption{\centering
    HoloOcean Rough Terrain 3D MSE (cm$^2$, ↓) scores across optimization steps and learning rates.
    Evaluated using the generic viewpoint (GV) approach with \(\lambda_\text{TV}=0.1\).
    For each number of steps, the best learning rate is shown in bold. The overall optimal performance is achieved at 200 steps with a learning rate of \num{1e-4}.
  }
  \label{tab:steps_lr_ablation_mse}
  \begin{tabular}{@{}lccccc@{}}
    \toprule
    Steps / LR &
    $1\times10^{-6}$ &
    $1\times10^{-5}$ &
    $5\times10^{-5}$ &
    $1\times10^{-4}$ &
    $5\times10^{-4}$ \\
    \midrule
    50  & 7.034 & 6.423 & 5.169 & \textbf{4.834} & 5.984 \\ 
    75  & 6.998 & 6.147 & 4.808 & \textbf{4.529} & 6.160 \\ 
    100 & 6.960 & 5.916 & 4.595 & \textbf{4.364} & 6.750 \\ 
    150 & 6.886 & 5.556 & 4.370 & \textbf{4.219} & 9.056 \\ 
    200 & 6.813 & 5.292 & 4.279 & \textit{\textbf{4.200}} & 12.398 \\ 
    300 & 6.673 & 4.938 & \textbf{4.286} & 4.325 & 21.012 \\ 
    400 & 6.540 & 4.721 & \textbf{4.399} & 4.568 & 30.553 \\ 
    \bottomrule
  \end{tabular}
\end{table*}

As shown in Tab.~\ref{tab:steps_lr_ablation_mse}, very small learning rates converge slowly and leave significant error even after many steps, while very large learning rates overshoot and degrade performance. The optimization remains stable up to a learning rate around \(5\times10^{-4}\), where it begins to create extreme geometry resulting in large error. A mid–range value around \(10^{-4}\) provides a stable compromise. It typically achieves optimal accuracy around \num{150}-\num{200} steps.

\subsection{Generic Viewpoint}
\label{sec:gv_ablate}

The generic–viewpoint prior is implemented by sampling base–plane orientations that cover a specified fraction of the sonar elevation span. At each optimization step, we sample a random plane within the specified range (from min. coverage to full coverage) and apply the corresponding tilt to the latent seafloor. This encourages reconstructions that remain consistent under small changes to the supporting plane.

To test how this prior performs under different sampling ranges, we vary the minimum coverage of the elevation span and evaluate the 3D MSE on all three datasets. Low coverage allows shallow planes that only intersect part of the range. High coverage forces the plane to span most of the elevation arc, which  restricts the feasible orientations.

\begin{table*}[h]
  \centering
  \caption{\centering
    {3D MSE (cm$^2$, ↓) across different generic viewpoint sampling schemes.}
    Evaluated with \(\lambda_\text{TV}=0.1\).
  }
  \label{tab:gv_ablate}
  \begin{tabular}{@{}lccc@{}}
    \toprule
    GV Min. Coverage &
    In-Distribution &
    HoloOcean &
    HO Rough \\
    \midrule
    0\%  & 0.821 & 1.036 & 4.958 \\
    10\% & 0.792 & 0.963 & 4.633 \\
    20\% & 0.683 & 0.908 & 4.325 \\
    30\% & 0.632 & 0.890 & 4.223 \\
    40\% & 0.601 & \textbf{0.885} & \textbf{4.165} \\
    50\% & 0.583 & 0.894 & 4.174 \\
    60\% & 0.578 & 0.907 & 4.213 \\
    70\% & \textbf{0.577} & 0.917 & 4.285 \\
    80\% & 0.581 & 0.933 & 4.384 \\
    90\% & 0.599 & 0.960 & 4.532 \\
    \bottomrule
  \end{tabular}
\end{table*}

As shown in Tab.~\ref{tab:gv_ablate}, the generic viewpoint prior remains stable across all sampling schemes. However, it clearly benefits from limiting very shallow planes, likely due to the sparse sampling and extreme geometry which occurs in this zone. Additionally, near very high coverage restrictions (around 90\%) the model performs similarly to the high-tilt approach, since we are no longer sampling over a broad range of small tilt variations. The best performance in-distribution occurs when we sample from a minimum coverage of 70\%, and for both HoloOcean datasets the best performance occurs when the minimum coverage is 40\%.

\section{Runtime and Scaling}
\label{sec:runtime_memory}

We profile the runtime and GPU memory usage of our differentiable sonar renderer and single-frame inversion on a desktop RTX~5090 GPU. We vary the number of range and azimuth bins around typical imaging sonar settings (Range $\in \{300, 400, 500, 600\}$, Azimuth $\in \{48, 96, 128\}$). For each configuration we report the average forward render time, the wall-clock time required to complete a full optimization of a single frame with 150 gradient steps, and the peak GPU memory recorded by PyTorch during the run.

We provide results at two elevation sampling levels. In the first level we cast 1500 elevation rays per azimuth beam. This configuration prioritizes speed and is sufficient for most of our experiments. In the second level we use 3000 rays per beam, which provides higher fidelity results, particularly when the number of range bins is large. Runtime and memory scale approximately linearly with the number of azimuth beams, the number of range bins, and the number of elevation rays.

At a realistic ARIS-like resolution of $96 \times 500$ bins, the 1500-ray rendering requires 83~ms per forward render, 12.5~s for a full 150-step reconstruction, and a peak of 9.7~GiB of GPU memory. With 3000 rays the same configuration requires 165~ms per render, 24.8~s for reconstruction, and 19.4~GiB of peak memory. The largest configuration with 3000 rays and $600 \times 128$ bins exceeds 32~GiB and does not fit in GPU memory.

\begin{table*}[]
  \centering
  \caption{\centering
    {Runtime and peak GPU memory for 1500 and 3000 elevation rays per azimuth beam on an RTX~5090 GPU}. We report average forward render time, total optimization time for 150 steps, and peak GPU memory.
    The configuration with 3000 elevation rays, 600 range bins, and 128 azimuth bins exceeds 30~GiB of memory and does not fit.}
  \label{tab:runtime_full}
  \begin{tabular}{@{}rccc@{\hspace{2em}}rccc@{}}
    \toprule
    \multicolumn{4}{c}{\(n_\mathrm{el}=1500\)} &
    \multicolumn{4}{c}{\(n_\mathrm{el}=3000\)} \\
    Range bins & Render (ms) & Optim. (s) & Peak VRAM (GiB) &
    Range bins & Render (ms) & Optim. (s) & Peak VRAM (GiB) \\
    \midrule
    \multicolumn{8}{@{}l}{\textbf{48 Azimuth Beams}} \\
    300 & 31  &  4.6 &  2.9 & 300 & 56  &  8.3 &  5.8 \\
    400 & 37  &  5.6 &  3.9 & 400 & 70  & 10.5 &  7.7 \\
    500 & 45  &  6.7 &  4.8 & 500 & 84  & 12.6 &  9.7 \\
    600 & 52  &  7.8 &  5.8 & 600 & 98  & 14.8 & 11.6 \\
    \midrule
    \multicolumn{8}{@{}l}{\textbf{96 Azimuth Beams}} \\
    300 & 57  &  8.6 &  5.8 & 300 & 107 & 16.1 & 11.6 \\
    400 & 71  & 10.6 &  7.7 & 400 & 134 & 20.1 & 15.5 \\
    500 & 83  & 12.5 &  9.7 & 500 & 165 & 24.8 & 19.4 \\
    600 & 100 & 15.0 & 11.6 & 600 & 197 & 29.5 & 23.2 \\
    \midrule
    \multicolumn{8}{@{}l}{\textbf{128 Azimuth Beams}} \\
    300 & 73  & 10.9 &  7.8 & 300 & 148 & 22.2 & 15.5 \\
    400 & 94  & 14.2 & 10.3 & 400 & 185 & 27.7 & 20.6 \\
    500 & 113 & 17.0 & 12.9 & 500 & 223 & 33.5 & 25.8 \\
    600 & 132 & 19.8 & 15.5 & 600 &  &   & $>$30 \\
    \bottomrule
  \end{tabular}
\end{table*}

\section{3D Reconstructions of Real Riverbeds}
\label{sec:3d_reconstructions}

\begin{figure*}[t]
    \centering
    \includegraphics[width=1.0\linewidth]{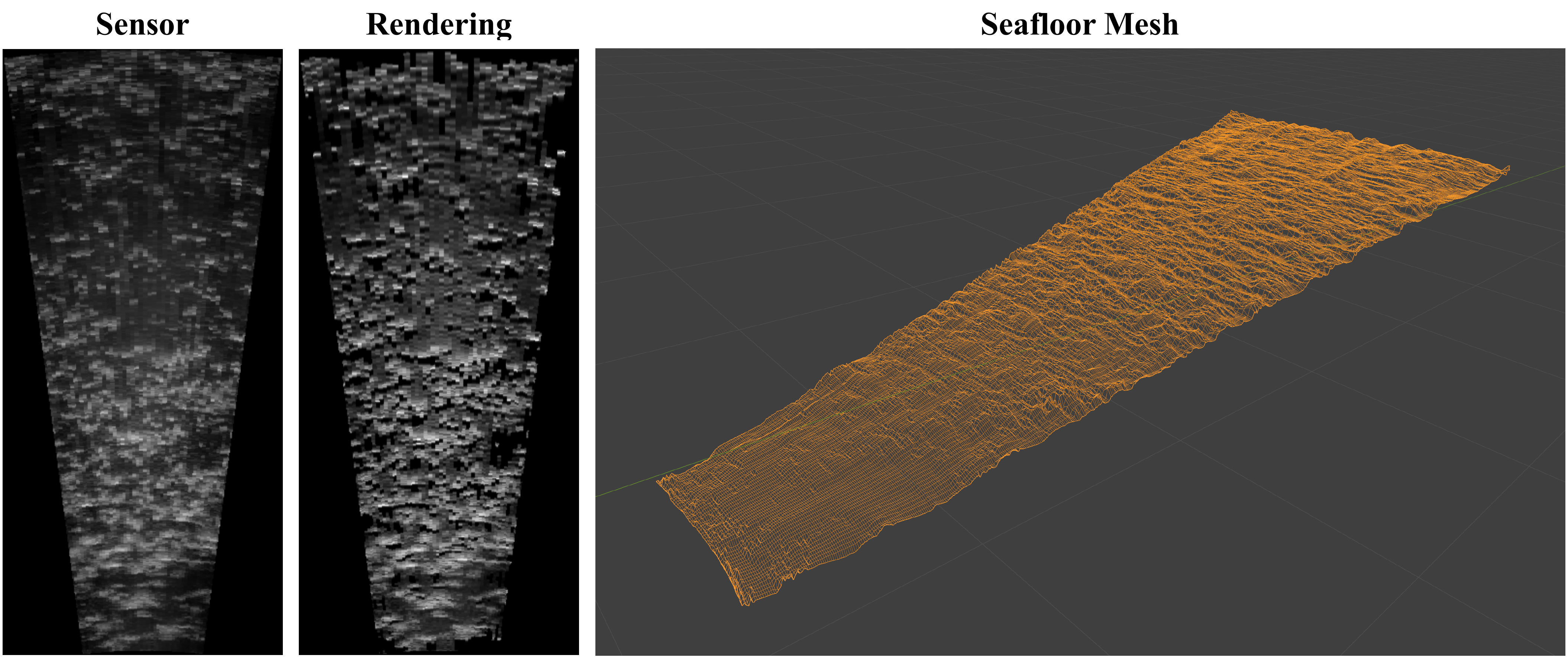}
    \caption{\centering {\bf Predicted seafloor mesh on Kenai Rightbank from a real sensor reading} (\textit{cf.} Sec.~\ref{sec:3d_reconstructions}).}
    \label{fig:riverA_fig1}
\end{figure*}

\begin{figure*}[t]
    \centering
    \includegraphics[width=1.0\linewidth]{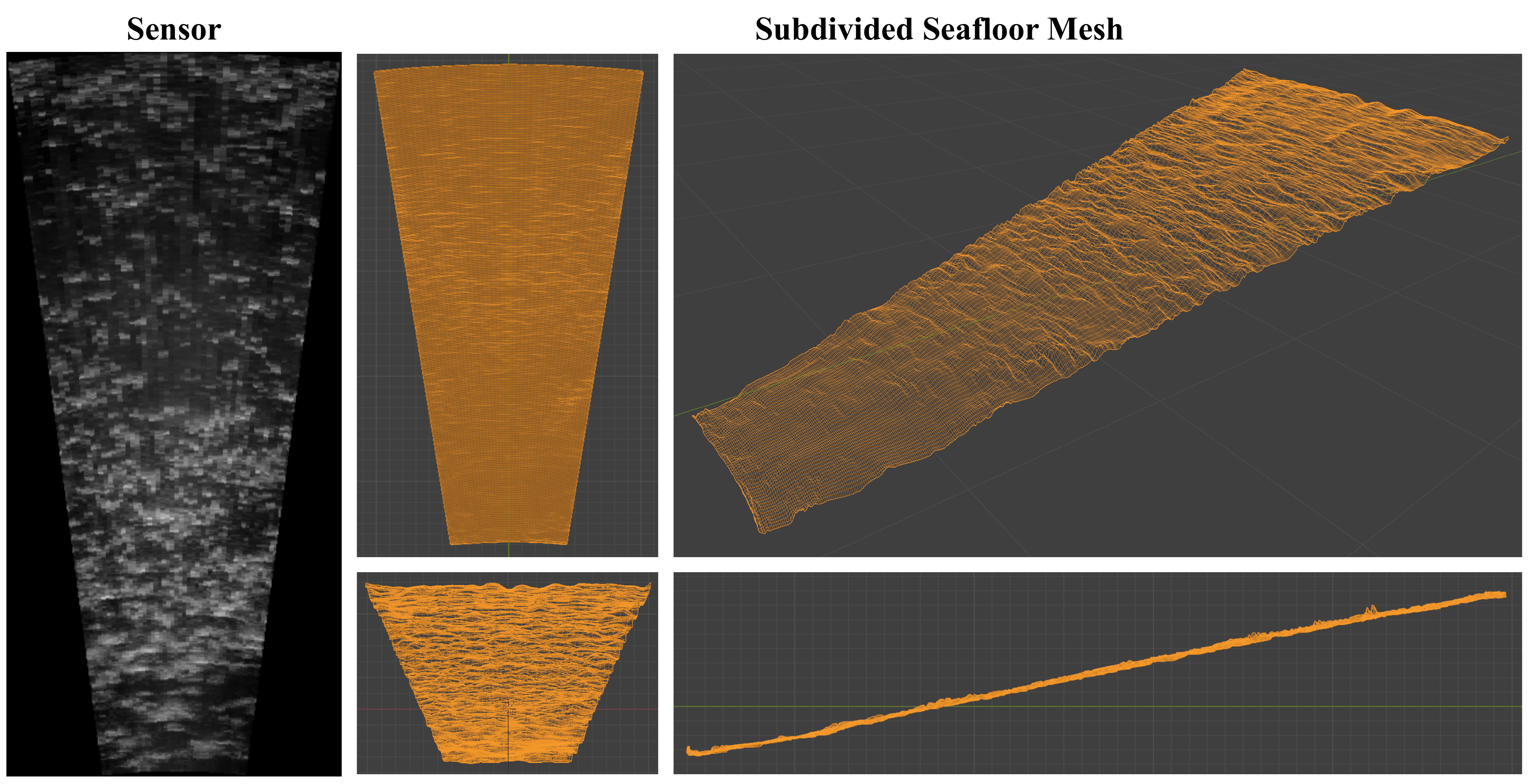}
    \caption{\centering {\bf Additional views of the recovered seafloor geometry on Kenai Rightbank} (\textit{cf.} Sec.~\ref{sec:3d_reconstructions}). On the left we show the target again, with additional orthographic views of the recovered height field on the right after one level of Catmull–Clark subdivision in Blender (applied only for visualization).}
    \label{fig:riverA_fig2}
\end{figure*}

\begin{figure*}[t]
    \centering
    \includegraphics[width=1.0\linewidth]{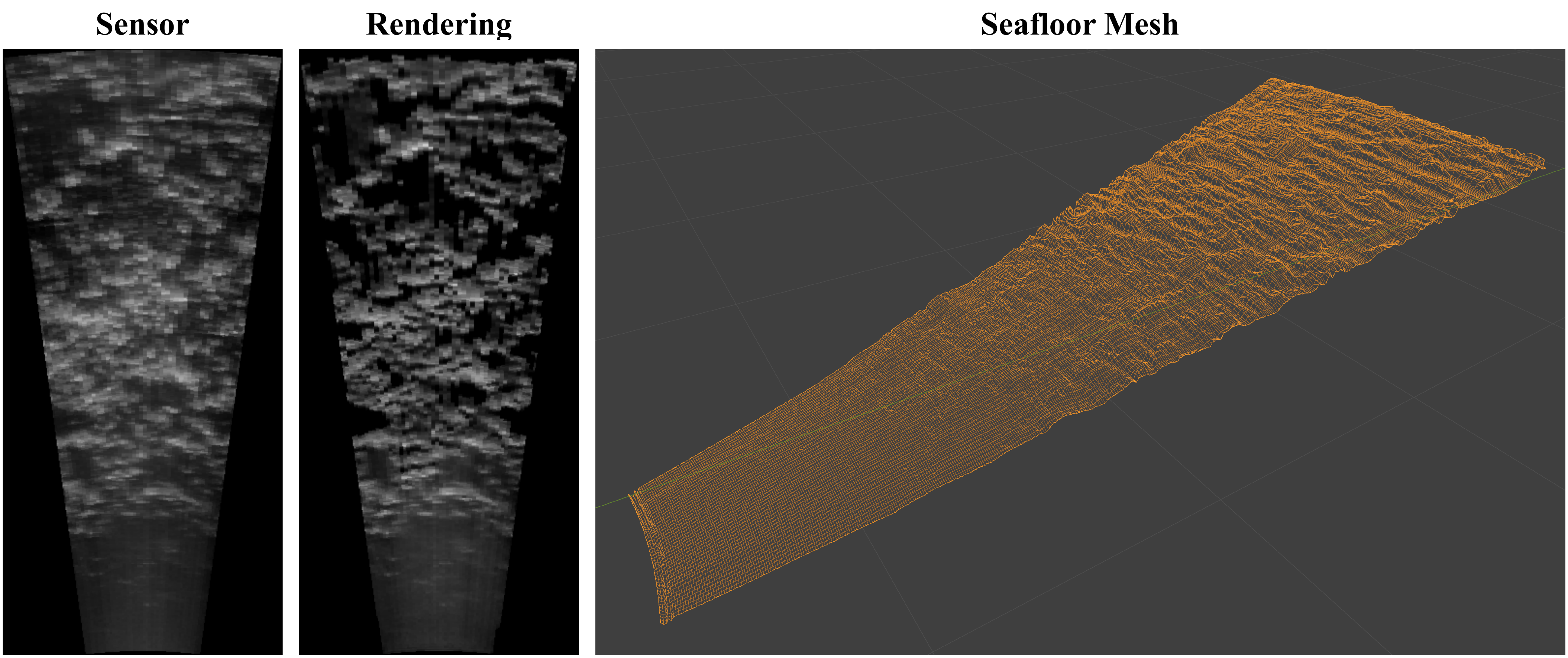}
    \caption{\centering {\bf Predicted seafloor mesh on Kenai Channel from a real sensor reading} (\textit{cf.} Sec.~\ref{sec:3d_reconstructions}).}
    \label{fig:riverB_fig1}
\end{figure*}

\begin{figure*}[t]
    \centering
    \includegraphics[width=1.0\linewidth]{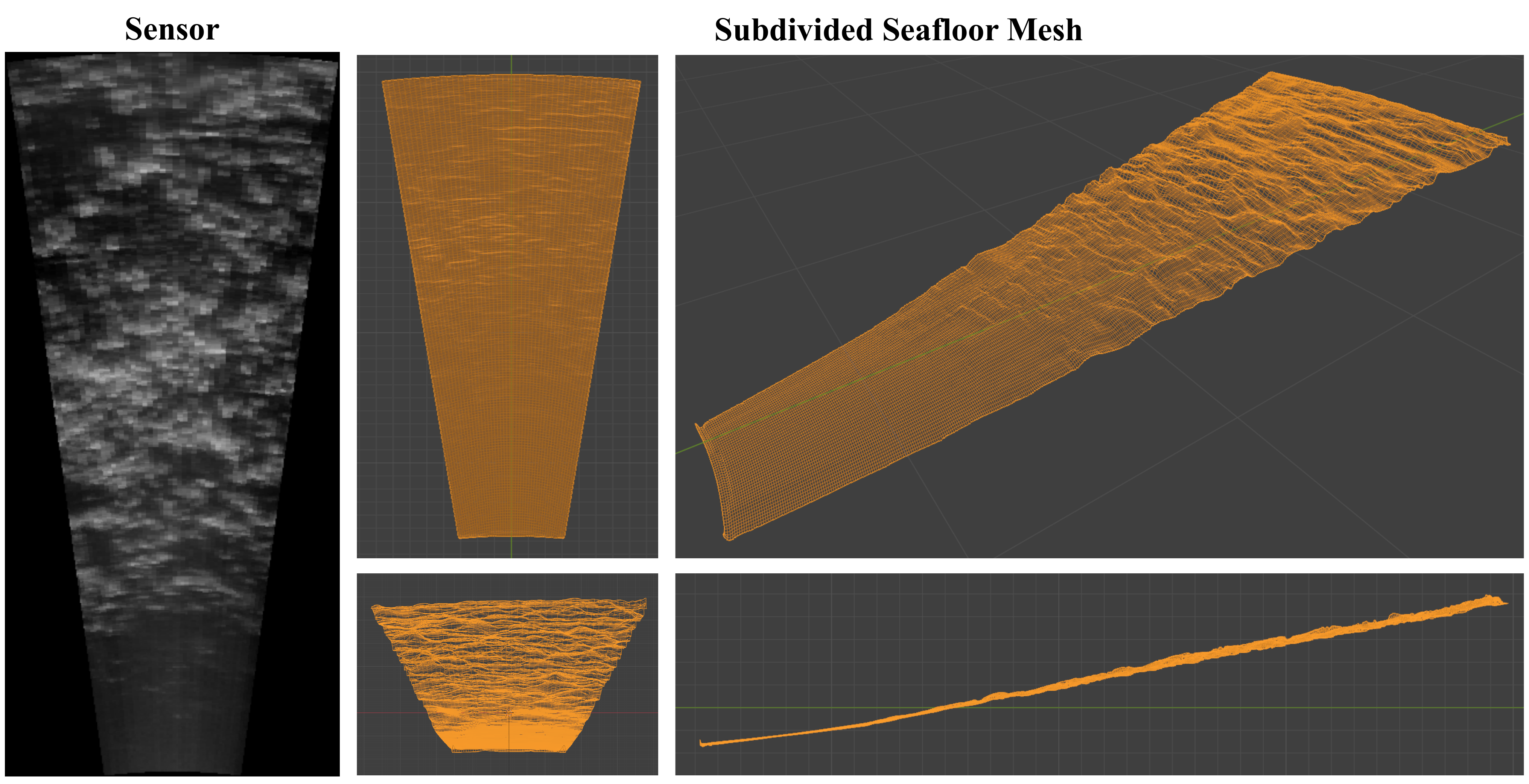}
    \caption{\centering {\bf Additional views of the recovered seafloor geometry on Kenai Channel} (\textit{cf.} Sec.~\ref{sec:3d_reconstructions}). On the left we show the target again, with additional orthographic views of the recovered height field on the right after one level of Catmull–Clark subdivision in Blender (applied only for visualization).}
    \label{fig:riverB_fig2}
\end{figure*}

In this section we present several examples of full 3D reconstructions produced by our differentiable renderer. These fits use a higher learning rate (\(3\times10^{-3}\)) and stronger TV regularization (\(\lambda_\text{TV}=0.5\)) than was used in the main paper. This choice increases contrast and occlusions in the rendered images, but also makes the underlying 3D structure easier to visualize. The full raw seafloor mesh for two real images (from Kenai Rightbank and Kenai Channel) are shown in~\cref{fig:riverA_fig1,fig:riverB_fig1} without modification. We show additional orthographic renders after applying one level of Catmull–Clark subdivision in Blender to smooth high–frequency noise in~\cref{fig:riverA_fig2,fig:riverB_fig2}.

\clearpage


\end{document}